\documentclass[runningheads]{llncs}

\usepackage[T1]{fontenc}
\usepackage{graphicx}
\usepackage{svg}
\usepackage{multirow}
\usepackage{booktabs}
\usepackage{epsfig}
\usepackage{amsmath}
\usepackage{amssymb}
\usepackage{tabularray}
\usepackage{colortbl}
\usepackage{subcaption}
\usepackage{orcidlink}
\usepackage{fontawesome}
\usepackage[symbol]{footmisc}
\usepackage{hyperref}

\usepackage{color}

\begin{document}

\title{Probing the Efficacy of Federated Parameter-Efficient Fine-Tuning of Vision Transformers for Medical Image Classification}
\titlerunning{Federated PEFT of ViT for Medical Image Classification}
%
\author{
Naif Alkhunaizi\textsuperscript{\dag}\orcidlink{0000-0002-7093-5034} \and 
Faris Almalik\textsuperscript{\dag}\orcidlink{0000-0002-7885-6285} \and 
Rouqaiah Al-Refai\orcidlink{}
\and \\
Muzammal Naseer \orcidlink{0000-0001-7663-7161} 
\and 
Karthik Nandakumar\orcidlink{0000-0002-6274-9725} \textsuperscript{(\faEnvelopeO)}}
\authorrunning{N. Alkhunaizi et al.}

\institute{Mohamed Bin Zayed University of Artificial Intelligence, Abu Dhabi, UAE
\email{\{naif.alkhunaizi, faris.almalik, rouqaiah.al-refai, muzammal.naseer,  karthik.nandakumar\}@mbzuai.ac.ae}}

\maketitle         

\begin{abstract}
With the advent of large pre-trained transformer models, fine-tuning these models for various downstream tasks is a critical problem. Paucity of training data, the existence of data silos, and stringent privacy constraints exacerbate this fine-tuning problem in the medical imaging domain, creating a strong need for algorithms that enable collaborative fine-tuning of pre-trained models. Moreover, the large size of these models necessitates the use of parameter-efficient fine-tuning (PEFT) to reduce the communication burden in federated learning. In this work, we systematically investigate various federated PEFT strategies for adapting a Vision Transformer (ViT) model (pre-trained on a large natural image dataset) for medical image classification. Apart from evaluating known PEFT techniques, we \textit{introduce new federated variants of PEFT algorithms} such as visual prompt tuning (VPT), low-rank decomposition of visual prompts, stochastic block attention fine-tuning, and hybrid PEFT methods like low-rank adaptation (LoRA)+VPT. Moreover, we perform a \textit{thorough empirical analysis} to identify the optimal PEFT method for the federated setting and understand the impact of data distribution on federated PEFT, especially for out-of-domain (OOD) and non-IID data. The key insight of this study is that while most federated PEFT methods work well for in-domain transfer, there is a substantial accuracy vs. efficiency trade-off when dealing with OOD and non-IID scenarios, which is commonly the case in medical imaging. Specifically, every order of magnitude reduction in fine-tuned/exchanged parameters can lead to a $4\%$ drop in accuracy.
Thus, the initial model choice is crucial for federated PEFT. It is preferable to use medical foundation models learned from in-domain medical image data (if available) rather than general vision models. Code will be provided upon acceptance.

\end{abstract}

\section{Introduction}
\label{sec:introduction}


Transformer models pre-trained on large-scale data can serve as a foundation for a wide range of downstream tasks \cite{bommasani2022opportunities}. While many general vision foundation models are available \cite{dosovitskiy2020image,CLIP}, developing generic medical foundation models is a challenge due to the diversity of imaging modalities and limited availability of well-annotated data \cite{ZhangMedicalFoundationModels2024}. Consider the scenario where a healthcare organization wants to learn transformer models for a range of medical image classification tasks such as chest x-ray disease classification \cite{chestxray_2,chestxray_1}, melanoma classification \cite{skincancer_2,skincancer_1}, and tumor categorization \cite{brain_2,brain_1}. There are two main challenges in this problem setting. Firstly, the organization may not have sufficient training data for each task to learn task-specific models from scratch. This can be addressed by fine-tuning a model that is pre-trained on a large-scale, independent dataset (transfer learning) for the task(s) at hand \cite{transfer_learning}. Secondly, storing a separate model for each task is inefficient due to the large size of transformer models. Parameter-efficient fine-tuning (PEFT) methods such as subset fine-tuning \cite{three_things}, adapter \cite{AdaptFormer}, low-rank adaptation (LoRA) \cite{lora}, and prompt tuning \cite{VPT} can mitigate this problem by fine-tuning only a small number of parameters for each task and storing the base model along with minimal task-specific parameters. Most PEFT methods exploit the inherently modular transformer architecture (characterized by a sequence of identical self-attention blocks processing a set of tokens).

In some medical imaging applications, even fine-tuning of pre-trained models may not be feasible when a hospital has data only from a few patients. However, a consortium of hospitals may be willing to collaborate to realize PEFT. This introduces the additional challenge of privacy because healthcare data is often regulated by strict privacy guidelines (e.g., GDPR, HIPAA), and it is not possible to pool data from multiple healthcare institutions centrally to enable machine learning. Federated learning (FL) \cite{mcmahan2017communication} can enable multiple entities to train a model collaboratively without sharing raw data. However, regular exchange of parameters in FL can become a communication burden, especially if the models are large. Hence, the combination of FL and PEFT is an ideal solution that can effectively solve multiple issues (paucity of data, storage of multiple large models, communication cost, and data privacy) simultaneously \cite{zhuang2023foundation}.

In this work, we consider a Vision Transformer (ViT) \cite{dosovitskiy2020image} model pre-trained on natural images as an illustrative example and explore federated PEFT in a cross-silo setting (with a small number of clients), aiming to answer the following questions: \textbf{(i)} Which PEFT method works well in conjunction with FL and provides the best accuracy vs. efficiency trade-off? \textbf{(ii)} Can federated PEFT transfer well for out-of-domain (OOD) and non-IID (independent, identically distributed) data encountered in medical image analysis? To the best of our knowledge, this is the first study that attempts to systematically study various PEFT strategies for ViTs within the FL framework. Our main contributions are:
\begin{enumerate}
    \item \textbf{New federated variants of PEFT methods}: We are the first to investigate visual prompt tuning (VPT) and low-rank decomposition of visual prompts (DVPT) in a federated setting. We also introduce a new federated subset fine-tuning approach called stochastic block attention (SBA). Finally, we also consider hybrid methods such as combining LoRA with VPT.
    
    \item \textbf{Analysis of federated PEFT methods}: We demonstrate that there is indeed a substantial trade-off between parameter efficiency and model accuracy in federated PEFT, especially for \textit{out-of-domain tasks with non-IID client distributions}. Hence, one must proceed with caution when adapting a general vision model for medical image classification using federated PEFT.
   
\end{enumerate}

\section{Background and Related Work}
\label{sec:background}
\begin{figure}[t!]
    \centering    
    \includegraphics[width=0.72\textwidth]{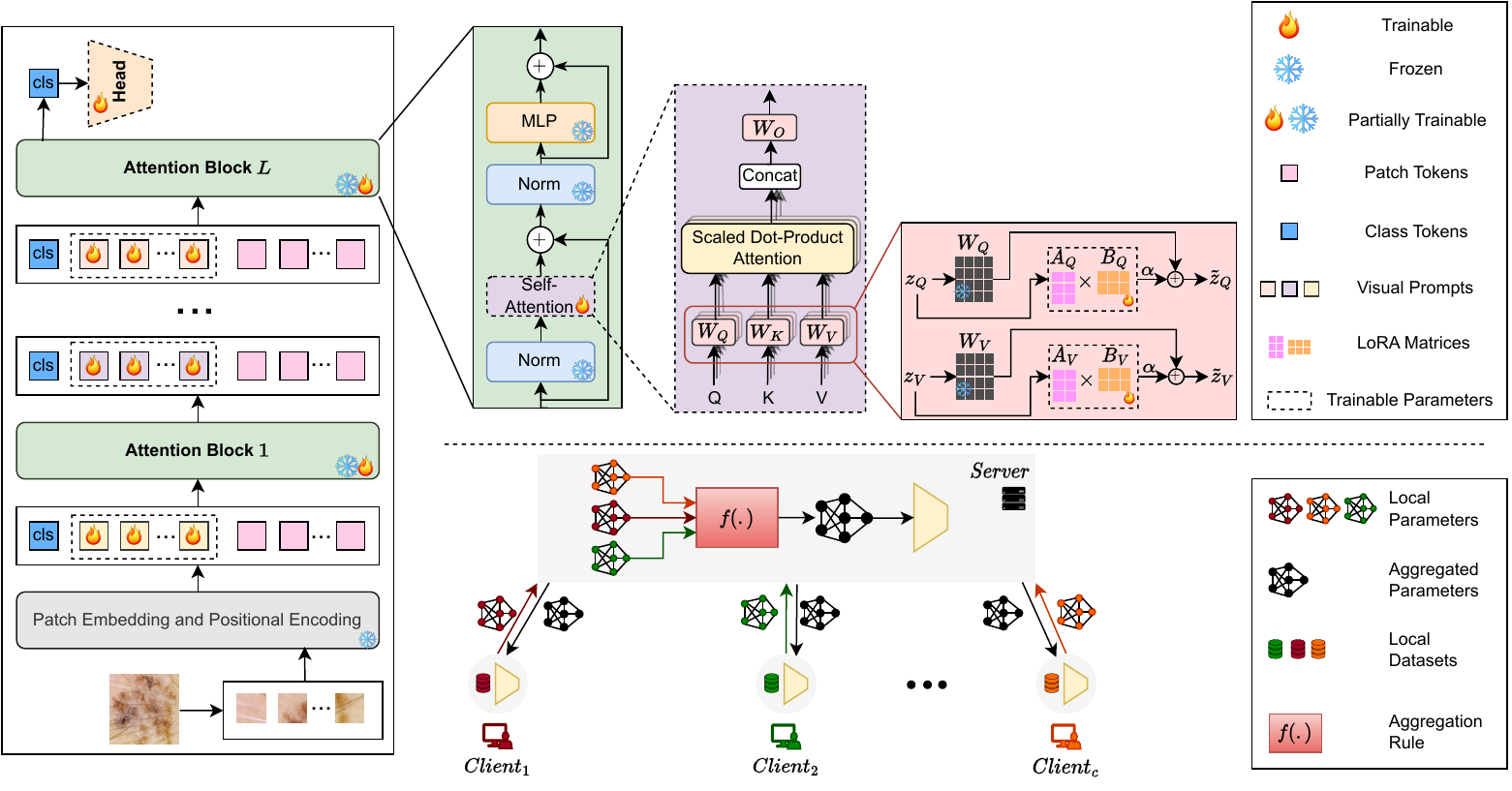}
    \caption{
    Adaptation of Vision Transformer (ViT) model using federated PEFT methods. Only the parameters marked as \textit{trainable} are exchanged between the clients and the server, while the frozen parameters are not communicated. 
    }
    \label{fig:method_figure}
\end{figure}

\noindent \textbf{Vision Transformer (ViT)}: A pre-trained ViT \cite{dosovitskiy2020image} can be considered as a feature extractor $\mathcal{V}_{\Psi}$ that maps a given input image $\mathbf{x}$ into a $d$-dimensional feature vector $\mathbf{f} \in \mathbb{R}^d$, where $\Psi$ denotes the complete set of ViT parameters. For image classification, a classification head $\mathcal{H}_{\eta}$ is typically trained to learn the mapping between $\mathbf{f}$ and the class label $y \in \{1,2,\cdots,K\}$, where $K$ is the number of classes and $\eta$ represents the parameters of the head $\mathcal{H}$. A ViT divides the  $\mathbf{x}$ into $S$ non-overlapping patches and a linear patch embedding layer $\mathcal{E}_{\Lambda}$ (with parameters $\Lambda$) is used to project each patch into $\mathbb{R}^d$, resulting in $\mathcal{T}_0 =\{\mathbf{t}_1,\cdots,\mathbf{t}_S\}$ \textit{patch tokens}. Additionally, a learnable \textit{class token} ($\tilde{\mathbf{t}}_0 \in \mathbb{R}^d$) is prepended to the sequence of patch tokens to assimilate the information as the tokens pass through $L$ transformer blocks (denoted by $\mathcal{G}_{\psi_\ell}$). The operations of each transformer block in a ViT can be represented as $\{\tilde{\mathbf{t}}_{\ell}, \mathcal{T}_{\ell}\} = \mathcal{G}_{\psi_{\ell}}(\{\tilde{\mathbf{t}}_{\ell-1}, \mathcal{T}_{\ell-1}\})$, $\ell  \in [1,L]$. The class token output by the $L^{\text{th}}$ (last) block (i.e., $\tilde{\mathbf{t}}_{L}$) can be considered as the final feature representation $\mathbf{f}$. Each transformer block, in turn, consists of three types of parameters (Fig. \ref{fig:method_figure}) - $\phi_{\ell}$ denotes the layer normalization parameters of the $\ell^{\text{th}}$ block, $\theta_{\ell}$ denotes the weight matrices of the multi-head self-attention (MHSA) layer of the $\ell^{\text{th}}$ block, and $\omega_{\ell}$ represents the parameters of the multi-layer perceptron (MLP) of the $\ell^{\text{th}}$ block. For convenience, let $\Phi = \{\phi_{\ell}\}_{\ell=1}^{L}$, $\Theta = \{\theta_{\ell}\}_{\ell=1}^{L}$, and $\Omega = \{\omega_{\ell}\}_{\ell=1}^{L}$ denote the collection of normalization, MHSA, and MLP parameters of all the $L$ blocks, respectively. Similarly, $\psi_l = \{\phi_\ell,\theta_\ell,\omega_\ell\}$ denote the set of all parameters of the $\ell^{\text{th}}$ block. Thus, ViT parameters can be summarized as $\Psi = \{\Lambda,\Phi,\Theta,\Omega\} = \{\Lambda,\psi_1,\cdots,\psi_L\}$ and ViT operations can be summarized as $\tilde{\mathbf{t}}_{L} = \mathcal{V}_{\Psi}(\mathbf{x}) = \mathcal{G}_{\psi_L}(\cdots(\mathcal{G}_{\psi_1}(\{\tilde{\mathbf{t}}_0, \mathcal{E}_{\Lambda}(\mathbf{x})\})))$. Since training a ViT from scratch requires a large dataset due to the lack of inductive bias \cite{inductive_bias_1}, fine-tuning has been the de-facto approach to adapt pre-trained ViTs for downstream tasks \cite{vit_object_detection_1}. 

\noindent \textbf{Parameter-Efficient Fine-Tuning (PEFT)}: PEFT methods achieve efficient adaptation of large pre-trained models \cite{peft3,lora,peft2} by learning only a limited number of parameters. \textit{Linear probing} learns only the head parameters ${\eta}$ and it represents the lower bound for all PEFT methods. In contrast, \textit{full fine-tuning} involves updating all the ViT parameters ($\Psi$) in addition to the head ($\eta$). \textit{Subset fine-tuning} methods fine-tune only a chosen subset of the pre-trained model parameters such as the last few layers of the network (e.g., \cite{k_layers1}) or the MHSA layer within each ViT block (e.g., \cite{three_things}). \textit{Visual Prompt Tuning (VPT)} \cite{VPT} introduces a set of $R$ learnable visual prompts before each ViT block, represented by $\mathcal{P}_{v} = \{\mathcal{P}_{v_\ell}\}_{\ell=1}^L \in \mathbb{R}^{L \times R \times d}$. The operations of each transformer block in a visually prompted ViT can be represented as $\{\tilde{\mathbf{t}}_{\ell}, \underline{\hspace{0.2cm}},\mathcal{T}_{\ell}\} = \mathcal{G}_{\psi_{\ell}}(\{\tilde{\mathbf{t}}_{\ell-1},\mathcal{P}_{v_\ell}, \mathcal{T}_{\ell-1}\})$, $\ell \in [1,L]$. During fine-tuning, only the prompts $\mathcal{P}_{v}$ are updated and the ViT parameters are unchanged. \textit{Low-Rank Adaptation (LoRA)} \cite{lora} injects trainable low-rank matrices in parallel to the attention layer \cite{peft3} while keeping the pre-trained model weights frozen. The MHSA parameters of a block $\ell$ can be considered as a collection of four weight matrices denoted as $\theta_{\ell} = \{\mathbf{W}_{O,\ell},\mathbf{W}_{Q,\ell},\mathbf{W}_{K,\ell},\mathbf{W}_{V,\ell}\}$. In LoRA, the updates to $\mathbf{W}_{Q,\ell}$ and $\mathbf{W}_{V,\ell}$ are decomposed into a pair of low rank matrices $\mathbf{A} \in \mathbb{R}^{r \times d}$ and $\mathbf{B} \in \mathbb{R}^{d \times r}$, where $r$ represents the rank of the two matrices. Let $z_{*,\ell}$ and $\tilde{z}_{*,\ell}$ be the input and output, respectively, of an attention layer in the $\ell^{\text{th}}$ block. Then, LoRA operations can be summarized as:

\begin{equation}
    \label{eq:processing_lora}
    \begin{split}
        \tilde{z}_{Q, \ell} &= \mathbf{W}_{Q,\ell}  z_{Q, \ell} + \alpha \mathbf{B}_{Q,\ell}  \mathbf{A}_{Q,\ell}   z_{Q, \ell}, \\ 
         \tilde{z}_{V, \ell} &= \mathbf{W}_{V,\ell}  z_{V, \ell} + \alpha  \mathbf{B}_{V,\ell}  \mathbf{A}_{V,\ell}   z_{V, \ell}.
    \end{split}
\end{equation} 

\noindent Here, $\mathcal{A} = \{\mathbf{A}_{Q,\ell},\mathbf{A}_{V,\ell}\}_{\ell=1}^L$ and $\mathcal{B} = \{\mathbf{B}_{Q,\ell},\mathbf{B}_{V,\ell}\}_{\ell=1}^L$ are the only learnable parameters, and $\alpha$ is a fixed scalar. Recently, PEFT methods have also been studied in the FL context. While textual prompt learning via FL was proposed in \cite{guo2023promptfl,prompts2}, a FL extension to LoRA was proposed in \cite{slora}.

\section{Federated Parameter-Efficient Fine-Tuning Methods}
\label{sec:method}

\noindent\textbf{Problem Statement}: We assume that a ViT feature extractor $\mathcal{V}_{\Psi_0}$ that is already pre-trained on a large independent dataset is available at the server. The goal of the server is to collaborate with the $C$ clients to fine-tune the pre-trained ViT feature extractor $\mathcal{V}_{\Psi_0}$ and learn the task-specific classification head $\mathcal{H}_{\eta}$ in a federated fashion while \textit{maximizing task-specific performance} and \textit{minimizing the number of parameters that are tuned and exchanged}. The server initializes $\eta$ as $\eta_0$ and broadcasts both $\Psi_0$ and $\eta_0$ to all the clients before the collaboration begins. At the end of $T$ collaboration rounds, the objective is to obtain $\Psi_T$ and $\eta_T$, which are fine-tuned for the specified task. By minimizing the number of parameters that are tuned and exchanged, we seek to reduce both the communication costs between the clients and the server as well as the memory footprint required to store the task-specific parameters.

\noindent\textbf{Vanilla Federated Learning (FedAvg)}: \cite{mcmahan2017communication} Given an appropriate per-client loss function $\mathcal{L}^{(c)}(\Psi,\eta)$, the global loss function is defined as:

\begin{equation}
\label{eq:global_loss_func}
    \mathcal{L}(\Psi,\eta) = \sum_{c=1}^{C} \frac{N^{(c)}}{N} \mathcal{L}^{(c)}(\Psi,\eta), 
\end{equation}

\noindent where $N = \sum_{c=1}^{C} N^{(c)}$ and $N^{(c)}$ is the number of training samples available at client $c \in [1,C]$. Starting from ($\Psi_0$, $\eta_0$), $\mathcal{L}(\Psi,\eta)$ is iteratively minimized over $T$ collaboration rounds. At the start of round $t$, client parameters are initialized as: $\Psi_{t-1}^{(c)} = \Psi_{t-1}$ and $\eta_{t-1}^{(c)} = \eta_{t-1}$, $\forall~t \in [1,T]$. In round $t$, the clients obtain:

\begin{equation}
    \Psi_{t}^{(c)},\eta_{t}^{(c)} = \underset{\Psi,\eta}{\arg \min} ~ \mathcal{L}^{(c)}(\Psi,\eta).
\end{equation}

\noindent At the end of round $t$, the server aggregates the client parameters as:

\begin{equation}
    \Psi_{t} = \sum_{c=1}^{C} \frac{N^{(c)}}{N} \Psi_{t}^{(c)},~ 
    \eta_{t} = \sum_{c=1}^{C} \frac{N^{(c)}}{N} \eta_{t}^{(c)}.
\end{equation}

The above formulation can be considered as the federated version of full fine-tuning. For federated linear probing, only $\eta$ is updated and $\Psi_{T} = \Psi_{0}$.

\subsection{Proposed Variants of Federated PEFT Methods}

\noindent \textbf{Federated Subset Fine-tuning}: Inspired by \cite{three_things}, we unfreeze \textit{all} MHSA parameters across all the $L$ blocks and fine-tune $\Theta$ in a federated fashion. Henceforth, we refer to this method as \textbf{all blocks attention (ABA)} with $\{\Theta,\eta\}$ being the only trainable parameters. The optimization formulation for ABA is $\min_{\Theta,\eta} ~ \mathcal{L}(\Psi = \{\Lambda,\Phi,\Theta,\Omega\}, \eta)$.

In the ABA method, clients must fine-tune and communicate the parameters of $L$ MHSA layers, which is roughly a third of the parameters involved in full fine-tuning. To further improve parameter efficiency, we propose \textbf{stochastic block attention (SBA)}, which requires updating parameters of only a \textit{single} MHSA layer in each collaboration round. Specifically, the server randomly samples a block $\ell^*$ in each round, where $\ell^* \in [1, L]$, and unfreezes its corresponding MHSA weights $\theta_{\ell^*}$. Then, all clients learn $\{\theta_{\ell^*}, \eta\}$ collaboratively as $\min_{\theta_{l^*},\eta} ~ \mathcal{L}(\Psi, \eta)$.

The SBA method involves learning only a fraction $(1/L)$ of the ABA parameters in each round, resulting in better communication efficiency. However, SBA requires the same storage as ABA because all the MHSA layers get updated over different rounds. In both ABA and SBA, FedAvg is used for aggregation.

\begin{figure}[t!]
\centering
\begin{subfigure}{0.4\textwidth}
    \includegraphics[width=\textwidth]{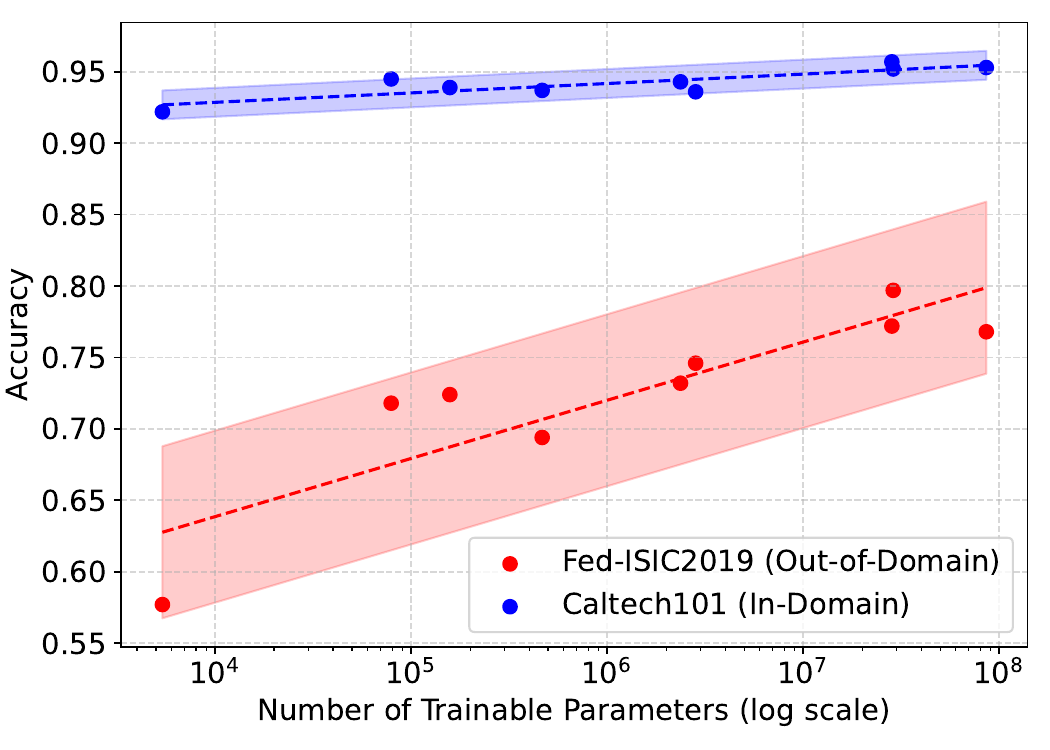}
    \caption{}
    \label{fig:tradeoff}
\end{subfigure}
\begin{subfigure}{0.58\textwidth}
    \includegraphics[width=\textwidth]{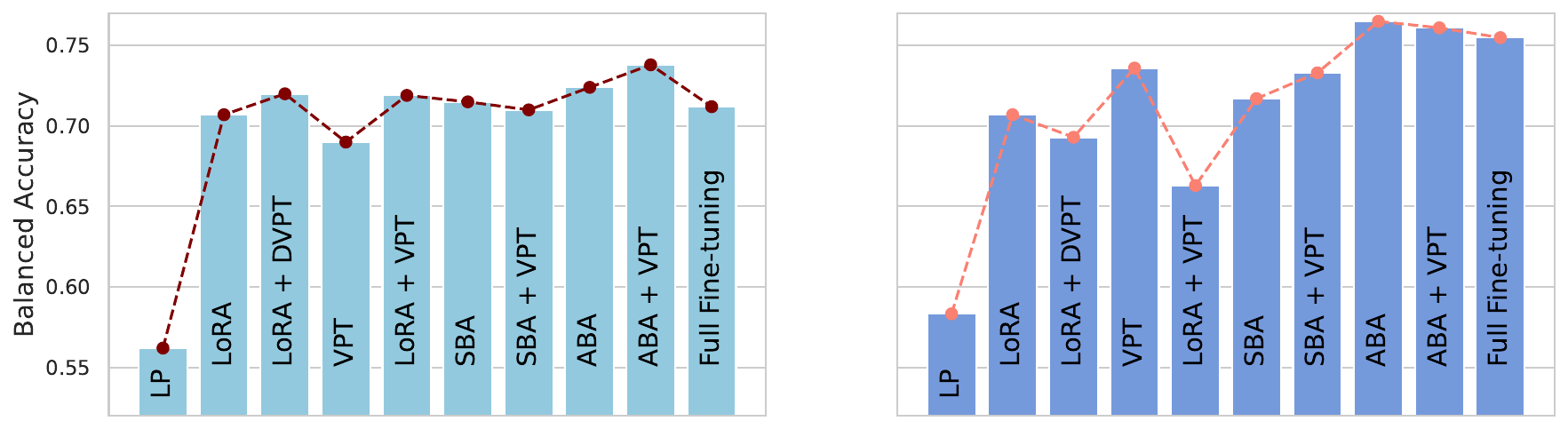}
    \caption{}
    \label{fig:fiveclients}
\end{subfigure}

\caption{(a) Accuracy vs. efficiency trade-off of various federated PEFT methods (Full Fine-tuning to LoRa shown in Table \ref{tab:methods_num_params}). The trade-off is more pronounced for OOD transfer (Fed-ISIC2019) compared to in-domain transfer (CalTech101). (b) Accuracy of federated PEFT methods on Fed-ISIC2019 with only $5$ clients (excluding client $4$), when (\textbf{Left}) base model is fine-tuned first with in-domain data (client $4$ data) and (\textbf{Right}) base model is pre-trained using natural images. Clearly, the in-domain base model shows less performance variability.}
\end{figure}

\noindent \textbf{Federated VPT}: When VPT is used, the augmented ViT parameters can be denoted as $\mathcal{V}_{[\Psi,\mathcal{P}_{v}]}$ and the objective is $\min_{\mathcal{P}_{v},\eta} ~ \mathcal{L}([\Psi,\mathcal{P}_{v}], \eta)$, where the visual prompts are again aggregated through FedAvg. To further reduce the number of exchanged parameters, clients can decompose the locally learned prompts \cite{DVPT} into low-rank matrices using singular value decomposition (SVD) \cite{SVD}. We refer to this technique as \textbf{Decomposed Visual Prompts (DVPT)}, where the prompts from all transformer blocks are concatenated to obtain a $(LR \times d)$ matrix, which is decomposed as $\mathcal{A}_{vp} \in \mathbb{R}^{LR \times r_v}$ and $\mathcal{B}_{vp} \in \mathbb{R}^{r_v \times d}$. Here, $r_v$ denotes the rank of visual prompt decomposition matrices $\mathcal{A}_{vp}$ and $\mathcal{B}_{vp}$. 

\noindent \textbf{Federated LoRA}: When LoRA is used, the augmented ViT parameters can be denoted as $\mathcal{V}_{[\Psi,\mathcal{A},\mathcal{B}]}$ and the objective function is $\min_{\mathcal{A}, \mathcal{B}, \eta} ~ \mathcal{L}([\Psi, \mathcal{A}, \mathcal{B}] , \eta)$. In federated LoRA, the server: (i) reconstructs back the weight update matrices sent by the clients as $\Delta \textbf{W}_{Q,\ell}^{(c)} = \mathbf{B}_{Q,\ell}^{(c)}\mathbf{A}_{Q,\ell}^{(c)}$ and $\Delta \textbf{W}_{V,\ell}^{(c)} = \mathbf{B}_{V,\ell}^{(c)}\mathbf{A}_{V,\ell}^{(c)}$, (ii) performs FedAvg based on these reconstructed matrices $\Delta \textbf{W}_{Q,\ell}^{(c)}$ and $\Delta \textbf{W}_{V,\ell}^{(c)}$, and (iii) applies SVD to the aggregated matrices to obtain the new global weight update matrices $\mathbf{B}_{Q,\ell}$, $\mathbf{A}_{Q,\ell}$, $\mathbf{B}_{V,\ell}$, and $\mathbf{A}_{V,\ell}$, which are sent back to the clients, where $\ell \in [1,L]$. Thus, federated LoRA provides communication efficiency for both the clients and the server, as well as involves less trainable parameters. The number of trainable parameters in LoRA is directly related to the rank $r$. Finally, another key area of investigation in this work is understanding the impact of integrating multiple PEFT methods in federated settings to adapt pre-trained ViTs.

\section{Results and Discussion}
\label{sec:Datasets and Implement}

\noindent\textbf{Datasets} : We conducted experiments on Fed-ISIC2019 \cite{terrail2022flamby}, HAM10000 \cite{HAM1000}, Caltech101 \cite{caltech101}, and Flowers102 \cite{flowers102} datasets. While the first two datasets are from the medical imaging domain (OOD), the latter two correspond to in-domain scenarios. Fed-ISIC2019 also has non-IID data distribution. 

\noindent \textbf{Implementation Setup}: We use the ImageNet \cite{ImageNet} pre-trained ViT-B/16 model from timm library \cite{timm}, with $L = 12$ blocks, $d = 768$, and $S = 196$ patches. We use normal distribution with $\mu = 0$ and $\sigma = 0.1$ for LoRA initialization, with $r = 4$ and $\alpha = 2$. For VPT, we set $R = 50$. For DVPT, we experimented with different rank values and set the rank $r_v$ to $8$. We run FL for $T=200$ collaboration rounds, employing an SGD optimizer with a learning rate of $10^{-2}$, and a batch size of $32$ using cross-entropy loss. We set the number of clients to $C = 6$ and allow parameter exchange in every round. All experiments were implemented using PyTorch $2.1.0$ and Nvidia A100 GPU. For more details on the datasets and experimental set-up, please refer to the supplementary material.

\begin{table}[t!]
\centering
\caption{Benchmarking of different approaches for federated fine-tuning of ViT. The number of exchangeable parameters (measured in \textit{millions}) associated with each method is highlighted. Each experiment was repeated three times using different seeds, with the table reporting the mean and standard deviation.}
\label{tab:methods_num_params}
\resizebox{0.97\textwidth}{!}{%
\begin{tabular}{lclcccccc} 
\toprule
 &  &  & \multicolumn{2}{c}{\textbf{Exchangeable~Parameters ($\downarrow$)~}} & \multicolumn{4}{c}{\textbf{Balanced Accuracy $(\uparrow)$}} \\ 
\cmidrule(l){6-9} \cmidrule{4-5}
\multicolumn{1}{c}{\textbf{Method}} & \textbf{Parameters} &  & \textbf{Number} & \textbf{Percentage ($\%$)} & \begin{tabular}[c]{@{}c@{}}\textbf{ HAM10000 }\\\textbf{ (IID) }\end{tabular} & \begin{tabular}[c]{@{}c@{}}\textbf{ Fed-ISIC2019 }\\\textbf{ (non-IID) }\end{tabular} & \begin{tabular}[c]{@{}c@{}}\textbf{ Caltech101 }\\\textbf{ (IID) }\end{tabular} & \begin{tabular}[c]{@{}c@{}}\textbf{ Flowers102 }\\\textbf{ (IID) }\end{tabular} \\ 
\cmidrule(lr){1-1}\cmidrule(lr){2-2}\cmidrule(lr){4-4}\cmidrule(lr){6-6}\cmidrule(lr){7-7}\cmidrule(lr){8-8}\cmidrule(lr){9-9}\cmidrule(lr){5-5}
Centralized & $\Psi$, $\eta$ &  & - & - & 0.805 $\pm$ 0.011 & 0.786 $\pm$ 0.015 & 0.964 $\pm$ 0.003 & 0.966 $\pm$ 0.006 \\ 
\cmidrule(l){1-2}
CentralizedVPT & $\mathcal{P}_v$, $\eta$ &  & - & - & 0.781 $\pm$ 0.008 & 0.746 $\pm$ 0.021 & 0.944 $\pm$ 0.002 & 0.966 $\pm$ 0.006 \\ 
\cmidrule(l){1-9}
{\cellcolor[rgb]{1,0.984,0.702}}Full Fine-tuning & {\cellcolor[rgb]{1,0.984,0.702}}$\Psi$, $\eta$ &  {\cellcolor[rgb]{1,0.984,0.702}}& {\cellcolor[rgb]{1,0.984,0.702}}86.0 & {\cellcolor[rgb]{1,0.984,0.702}}100 & {\cellcolor[rgb]{1,0.984,0.702}}0.791 ± 0.025 & {\cellcolor[rgb]{1,0.984,0.702}}0.768 $\pm$ 0.046 & {\cellcolor[rgb]{1,0.984,0.702}}0.956 $\pm$ 0.004 & {\cellcolor[rgb]{1,0.984,0.702}}0.970 $\pm$ 0.004 \\ 
\cmidrule(l){1-2}
ABA $+$ VPT & $\Theta$, $\eta$, $\mathcal{P}_v$ &  & 28.815 & 33.5 & \textbf{0.812 $\pm$ 0.019} & \textbf{0.797 $\pm$ 0.001} & 0.946 $\pm$ 0.006 & 0.940 $\pm$ 0.002 \\ 
\cmidrule(l){1-2}
ABA & $\Theta$, $\eta$ &  & 28.354 & 32.97 & 0.812 $\pm$ 0.024 & 0.772 $\pm$ 0.013 & \textbf{0.958 $\pm$ 0.001} & \textbf{ 0.969 $\pm$ 0.001} \\ 
\cmidrule(l){1-2}
SBA $+$ VPT & $\theta_l^*$, $\eta$, $\mathcal{P}_v$ &  & 2.829 & 3.29 & 0.792 $\pm$ 0.016 & 0.746 $\pm$ 0.014 & 0.942 $\pm$ 0.009 & 0.938 $\pm$ 0.009 \\ 
\cmidrule(l){1-2}
SBA & $\theta_l^*$, $\eta$ &  & 2.368 & 2.75 & 0.782 $\pm$ 0.009 & 0.732 $\pm$ 0.009 & 0.948 $\pm$ 0.005 & 0.962 $\pm$ 0.002 \\ 
\cmidrule(l){1-2}
LoRA $+$ VPT & \begin{tabular}[c]{@{}c@{}}$\mathcal{A}$, $\mathcal{B}$, \\ $\mathcal{P}_v$, $\eta$\end{tabular} &  & 0.613 & 0.71 & 0.784 $\pm$ 0.018 & 0.729 $\pm$ 0.014 & 0.937 $\pm$ 0.006 & 0.947 $\pm$ 0.005 \\ 
\cmidrule(l){1-2}
VPT & $\mathcal{P}_v$, $\eta$ &  & 0.466 & 0.54 & 0.789 $\pm$ 0.002 & 0.694 $\pm$ 0.013 & 0.939 $\pm$ 0.002 & 0.939 $\pm$ 0.005 \\ 
\cmidrule(l){1-2}
LoRA $+$ DVPT & \begin{tabular}[c]{@{}c@{}}$\mathcal{A}$, $\mathcal{B}$, $\mathcal{A}_{vp}$, \\ $ \mathcal{B}_{vp}$, $\eta$\end{tabular} &  & 0.231 & 0.27 & 0.772 $\pm$ 0.022 & 0.724 $\pm$ 0.011 & 0.940 $\pm$ 0.004 & 0.946 $\pm$ 0.008 \\ 
\cmidrule(l){1-2}
LoRA & $\mathcal{A}$, $\mathcal{B}$, $\eta$ &  & 0.152 & 0.18 & 0.770 $\pm$ 0.011 & 0.718 $\pm$ 0.004 & 0.949 $\pm$ 0.003 & 0.963 $\pm$ 0.006 \\ 
\cmidrule(l){1-9}
PromptFL & $\mathcal{P}_t$ &  & 0.026 & 0.03 & 0.384 $\pm$0.005 & 0.389 $\pm$ 0.003 & 0.929 $\pm$ 0.002 & 0.814 $\pm$ 0.006 \\ 
\cmidrule(l){1-2}
Linear Probing & $\eta$ &  & 0.005 & 0.006 & 0.714 $\pm$ 0.012 & 0.577 $\pm$ 0.015 & 0.924 $\pm$ 0.003 & 0.929 $\pm$ 0.011 \\ 
\cmidrule(l){1-2}
Local & $\Psi, \eta$ &  & 0.0 & 0.0 & 0.674 $\pm$ 0.019 & 0.291 $\pm$ 0.028 & 0.607 $\pm$ 0.068 & 0.458 $\pm$ 0.009 \\
\bottomrule
\end{tabular}
}
\end{table}
\label{sec:results and discussion}
\noindent \textbf{Results}: Results are summarized in Table \ref{tab:methods_num_params}, where the methods are divided into three groups. The first two rows correspond to centralized training, where data from all clients gets pooled at one location (violating privacy constraints). This setting serves as a useful reference to understand the impact of FL. The middle group of methods exchange parameters related to the ViT in a federated setting. Among these methods, federated full fine-tuning of ViT (in yellow) is used as the baseline to assess the various PEFT methods. The third group (last three rows) corresponds to the case where no ViT parameters are exchanged.

\noindent \textbf{Which federated PEFT method provides the best accuracy vs. efficiency trade-off?}
While the performance of ABA is comparable to the baseline across all datasets, SBA achieves good performance only on IID datasets and has a noticeable degradation in Fed-ISIC2019. Due to the non-IID nature of Fed-ISIC2019, the stochastic block selected at each round might lead to divergence in the training process in certain rounds. A similar trend was observed with VPT, LoRA, and linear probing. \textit{While most federated PEFT methods work well for the IID scenario}, easily achieving up to three orders of magnitude decrease in the exchangeable parameters at a marginal cost to accuracy, \textit{they exhibit sub-optimal performance when there is statistical heterogeneity across clients}.

\noindent \textbf{Can federated PEFT transfer well for OOD tasks?} Our main finding is that there is a trade-off between parameter efficiency and model accuracy in federated PEFT. While this trade-off is \textit{marginal for in-domain tasks} (approximately $0.5\%$ decrease in accuracy for every order of magnitude reduction in the number of parameters fine-tuned/exchanged), this trade-off becomes \textit{substantial for out-of-domain tasks with non-IID client distributions} (approximately $4\%$ decrease in accuracy for every order of magnitude reduction as shown in Figure \ref{fig:tradeoff}). Therefore, ABA is the best approach for OOD transfer, though it has less parameter efficiency. While existing wisdom is that PEFT can be achieved without compromising on model accuracy \cite{exploring}, we have demonstrated that the above claim is true only for in-domain tasks. For further validation, we first fine-tune the pre-trained ViT on client $4$ of Fed-ISIC2019 and attempt to again fine-tune this new ``pre-trained'' model using the remaining $5$ clients in a federated manner. Note that after the first fine-tuning, the classification head is discarded, but the feature extraction model is already familiar with the medical imaging domain. So, the second federated PEFT stage can be considered as in-domain transfer. As depicted in Fig. \ref{fig:fiveclients} (Left), there is little difference among the federated PEFT methods in this scenario, proving that they perform equally well for in-domain transfer. However, when the original ``pre-trained'' model is plugged back and collaboratively fine-tuned with the same $5$ clients (excluding client $4$), we observe significant variability in the accuracy (Fig. \ref{fig:fiveclients} (Right)).

\noindent \textbf{Can a combination of PEFT methods further improve performance?} We experimented with different combinations of PEFT methods by using ABA, SBA, and LoRA in conjunction with VPT. Note that since both attention fine-tuning (ABA and SBA) and LoRA attempt to update the attention weights, it does not make sense to combine them. The results show that \textit{combining VPT with attention fine-tuning is beneficial for OOD transfer, while it hurts in-domain transfer}. This finding is confirmed by observing a similar trend when comparing the LoRA+VPT method with LoRA. Furthermore, combining LoRA with DVPT improves parameter efficiency by $3\times$ while yielding almost similar results. 

\noindent \textbf{Comparison with PromptFL:} Federated learning of \textit{text prompts} led to drastic performance degradation, particularly for OOD tasks (HAM10000 and Fed-ISIC2019), highlighting the relative superiority of federated VPT over PromptFL.

\section{Conclusion}
\label{sec:Conclusion}
This work probed the efficacy of various federated PEFT methods to adapt pre-trained vision transformers for medical image classification, focusing on achieving optimal performance while minimizing communication costs. Through extensive experimentation, we show that PEFT methods exhibit limited efficacy when applied to heterogeneous and out-of-domain datasets across participating clients. Hence, we recommend that \emph{it is preferable to start fine-tuning with in-domain medical foundation models} (if available), rather than models pre-trained on natural images. Our findings also highlight the robustness of visual prompts over text prompts, especially when the task does not involve natural images.


\bibliographystyle{splncs04}
\bibliography{Arxiv_Version}
\newpage
\appendix
\begin{center}
\textbf{\large Supplementary Material}
\end{center}
\begin{figure}[h!]
\centering
\includegraphics[width=\textwidth]{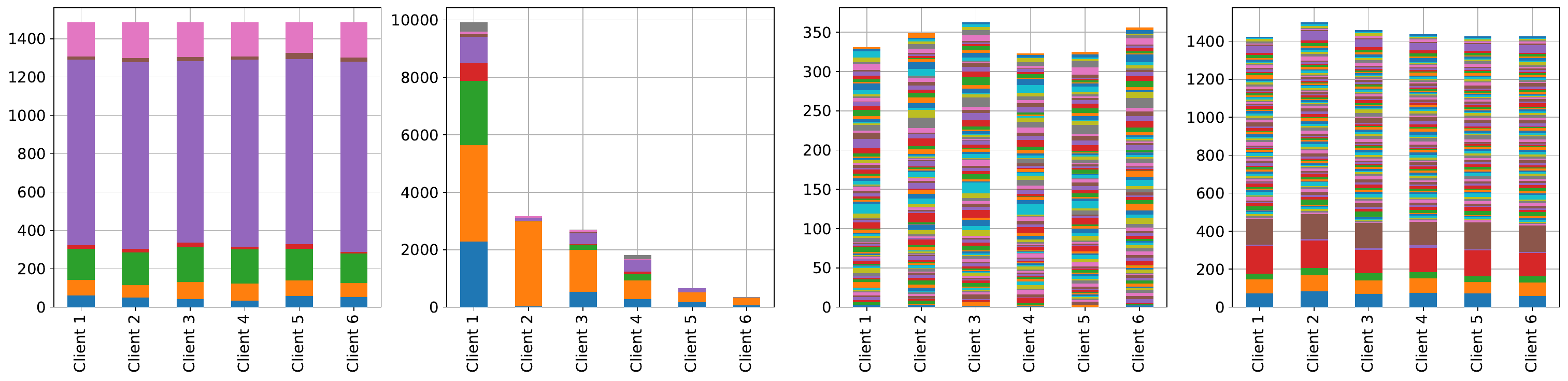}
\caption{From left to right, distribution of \textbf{HAM10000 (IID)}, \textbf{Fed-ISIC2019 (Non-IID)}, \textbf{Flowers102 (IID)}, and \textbf{Caltech101 (IID)} datasets. Each stacked bar represents the number of training samples, and each color represents a class. Fed-ISIC2019 \cite{terrail2022flamby} contains $23,247$ samples across eight melanoma classes. HAM10000 \cite{HAM1000} comprises $10,015$ dermoscopic images categorized into $7$ lesion types. We employ $80\%~(20\%)$ train (test) split for both these datasets. Caltech101 \cite{caltech101} has $101$ categories of natural images with a $50\%~(50\%)$ train (test) split. Flowers102 \cite{flowers102} includes 102 categories with a $25\% ~(75\%)$ train (test) split.}
\label{fig:distribution}
\end{figure}

\begin{figure}[h!]
\centering
\includegraphics[width=0.6\textwidth]{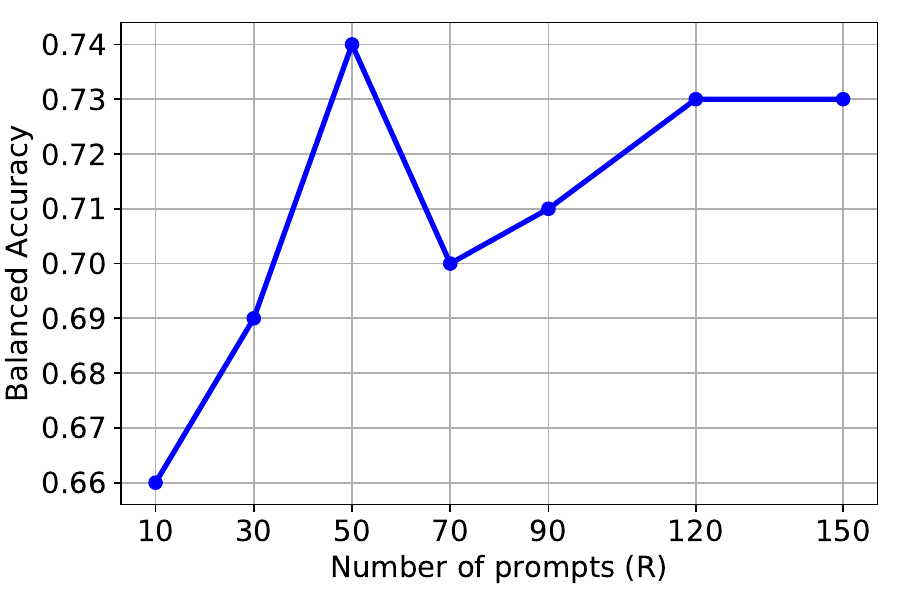}
\caption{Balanced accuracy with different number of prompts for the VPT method on Fed-ISIC2019 dataset. We found that optimal performance was achieved with $R = 50$ prompts.}
\label{fig:num_prompts}
\end{figure}
\newpage
\begin{table}[h!]
\centering
\caption{Balanced accuracy of LoRA with different initialization methods, scale, and rank for the HAM10000 and Fed-ISIC2019 datasets. Following \cite{lora}, we set all matrices in $\mathcal{B}$ to $\mathbf{0}$. We observe that initializing $\mathcal{A}$ based on a $Normal(0, 0.1)$ distribution with $r=4$ and $\alpha = 2$ represents the most effective trade-off between performance and the number of trained parameters associated with LoRA.}
\label{tab:initialization_types}
\renewcommand{\arraystretch}{0.9}
\resizebox{\textwidth}{!}{%
\begin{tabular}{@{}lclccccccccccccccccccc@{}}
\toprule
\multicolumn{2}{l}{} &
   &
  \multicolumn{3}{c}{Xavier} &
  \multicolumn{1}{l}{} &
  \multicolumn{3}{c}{Kaiming} &
  \multicolumn{1}{l}{} &
  \multicolumn{3}{c}{ImageNet} &
  \multicolumn{1}{l}{} &
  \multicolumn{3}{c}{\begin{tabular}[c]{@{}c@{}}Normal \\ $\mu = 0, \sigma= 0.5$\end{tabular}} &
  \multicolumn{1}{l}{} &
  \multicolumn{3}{c}{\begin{tabular}[c]{@{}c@{}}Normal \\ $\mu = 0, \sigma = 0.1$\end{tabular}} \\  \cmidrule(lr){4-6} \cmidrule(lr){8-10} \cmidrule(lr){12-14} \cmidrule(lr){16-18} \cmidrule(l){20-22} 
                             & Rank        &  & 4     &  & 8     &  & 4     &  & 8     &  & 4     &  & 8     &  & 4     &  & 8     &  & 4     &  & 8     \\ \cmidrule(lr){2-2} \cmidrule(lr){4-6} \cmidrule(lr){8-10} \cmidrule(lr){12-14} \cmidrule(lr){16-18} \cmidrule(l){20-22} 
\multirow{2}{*}{HAM10000}    & Scale = 2   &  & 0.816 &  & 0.816 &  & 0.810  &  & 0.798 &  & 0.799 &  & 0.804 &  & 0.646 &  & 0.536 &  & \textbf{0.826} &  & 0.801 \\
                             & Scale = 0.5 &  & 0.803 &  & 0.842 &  & 0.780 &  & 0.811 &  & 0.787 &  & 0.781 &  & 0.780 &  & 0.806 &  & 0.812 &  & 0.831 \\ \midrule
\multirow{2}{*}{FedISIC2019} & Scale = 2   &  & 0.754 &  & 0.777 &  & 0.765 &  & 0.785 &  & 0.757 &  & 0.773 &  & 0.495 &  & 0.572 &  & \textbf{0.757} &  & 0.786 \\
                             & Scale = 0.5 &  & 0.758 &  & 0.764 &  & 0.723 &  & 0.730 &  & 0.741 &  & 0.751 &  & 0.748 &  & 0.651 &  & 0.746 &  & 0.742 \\ \bottomrule
\end{tabular}%
}
\end{table}
\end{document}